\documentclass[camera-ready]{ecai} 

\usepackage{latexsym}
\usepackage{amssymb}
\usepackage{amsmath}
\usepackage{amsthm}
\usepackage{booktabs}
\usepackage{enumitem}
\usepackage{graphicx}
\usepackage{color}

\usepackage{algorithm}

\usepackage{latexsym}

\usepackage{pgfplots}
\pgfplotsset{compat=newest}
\usetikzlibrary{matrix}

\usepackage{subfig}
\usepackage{algorithm}
\usepackage{algpseudocode}
\usepackage{amsbsy}
\usepackage{amsmath}
\usepackage{amssymb}

\usepackage{multirow}
\usepackage{makecell}

\usepackage{color}
\usepackage{tabularray}

\newtheorem{definition}{Definition}

\newcommand{\BibTeX}{B\kern-.05em{\sc i\kern-.025em b}\kern-.08em\TeX}

\begin{document}


\begin{frontmatter}


\paperid{123} 


\title{Fact Probability Vector Based Goal Recognition}


\author[A]{\fnms{Nils}~\snm{Wilken}\orcid{0000-0003-1336-245X}\thanks{Corresponding Author. Email: nils.wilken@uni-mannheim.de.}}
\author[B]{\fnms{Lea}~\snm{Cohausz}}
\author[A]{\fnms{Christian}~\snm{Bartelt}}
\author[B]{\fnms{Heiner}~\snm{Stuckenschmidt}}

\address[A]{Institute for Enterprise Systems, University of Mannheim}
\address[B]{Data and Web Science Group, University of Mannheim}


\begin{abstract}
We present a new approach to goal recognition that involves comparing observed facts with their expected probabilities.
These probabilities depend on a specified goal $g$ and initial state $s_0$.
Our method maps these probabilities and observed facts into a real vector space to compute heuristic values for potential goals.
These values estimate the likelihood of a given goal being the true objective of the observed agent.
As obtaining exact expected probabilities for observed facts in an observation sequence is often practically infeasible, we propose and empirically validate a method for approximating these probabilities.
Our empirical results show that the proposed approach offers improved goal recognition precision compared to state-of-the-art techniques while reducing computational complexity.
\end{abstract}

\end{frontmatter}


\section{Introduction}
\label{sec:introduction}
Goal recognition is the task of recognizing the goal(s) of an observed agent given a sequence of actions executed or a sequence of states visited by the agent.
In this paper, we focus on observed action sequences for the theoretical discussion of the goal recognition problem.
This task is relevant in many application domains like crime detection \cite{geib2001plan}, pervasive computing \cite{wilken2021hybrid,geib2002problems}, or traffic monitoring \cite{pynadath1995accounting}.
Previous goal recognition systems often rely on the principle of Plan Recognition As Planning (PRAP) and, hence, utilize concepts and algorithms from the classical planning community to solve the goal recognition problem \cite{ramirez2009plan,ramirez2010probabilistic,sohrabi2016revisited,amado2018lstm}.
Nevertheless, a fundamental limitation of many systems from this area, which require computing entire plans to solve a goal recognition problem, is their computational complexity.
\citet{pereira2020landmark} approached this shortcoming by introducing a planning landmark-based heuristic approach, which outperforms existing goal recognition approaches and requires much less computation time.
However, as this approach only relies on fact landmarks, it neglects the information from all other observed facts.
In the context of classical planning, facts model the properties of the planning environment.
A planning state is defined by the subset of facts that hold true at that moment.

This paper presents a novel approach to goal recognition that involves comparing observed facts with their probability of being observed.
We define the probability of observing a fact $f$ as the probability of $f$ being part of an observation sequence (i.e., plan) that starts at the initial state $s_0$ and ends at a possible goal state $s_g$.
The exact meaning of a fact $f$ being part of an observation sequence is defined later in the paper by Definition \ref{def:observedFact}.
The proposed approach implements this comparison by mapping the fact observation probabilities per goal, the initial state $s_0$, and the currently observed state $s_t$ into a real vector space.
Based on these vector mappings, the proposed method computes a heuristic value for each possible goal based on these vector mappings.
The proposed heuristic estimates the probability that a possible goal $g$ is the actual goal $g^*$ of the observed agent.
We will refer to our proposed method as Fact Probability Vector Based Goal Recognition, or FPV, for short.

FPV and the planning landmark-based (PLR) method introduced by \citet{pereira2020landmark} fall in the same class of heuristic goal recognition approaches.
PLR uses the fact that a planning landmark $l$, by definition, has to be observed during an observation sequence that starts at $s_0$ and ends at $g$.
Based on this, PLR uses a heuristic that estimates the probability of a goal $g$ being the actual goal $g^*$ by counting how many planning landmarks out of all landmarks were already observed for $g$.
However, this limits the PLR approach to using only observed facts that are planning landmarks.
This problem becomes more severe when fewer planning landmarks exist in a planning domain.
One extreme case for such a domain is the grid-world domain, as we will discuss in Section \ref{sec:relaxedSimilarityRecognition}.
Only the initial state and the goal facts are fact landmarks in this domain.
To address this problem, FPV considers not solely the facts that are landmarks but \textit{all} facts that a given planning domain defines.

More explicitly, the contributions of this paper are:
\begin{itemize}
    \item In Section \ref{sec:relaxedSimilarityRecognition}, we propose a novel approach to goal recognition that involves comparing fact observation probabilities with actually observed facts.
    \item As the fact observation probabilities are usually not given, we propose a method to estimate the fact observation probabilities in Section \ref{sec:estimation}.
    \item In Section \ref{sec:evaluation}, we empirically show that FPV achieves better goal recognition precision, especially when dealing with low observability, and is more efficient regarding computation time than existing state-of-the-art methods.
\end{itemize}

\section{Background}
This section introduces relevant definitions in the context of classical planning and goal recognition.

\subsection{Classical Planning}
\label{subsec:classicalPlanning}
Classical planning uses a symbolic model of the planning domain that defines the properties of the planning environment (i.e., planning facts) and possible actions.
These actions are defined by their preconditions and effects.
Given an initial state and a goal, planning methods aim to construct an action sequence (i.e., plan) that transforms the initial state into a valid goal state (i.e., a state in which the goal description holds).
The generated plans might be strictly or approximately optimal depending on the planning method used.
For the theoretical analysis of FPV, we focus on the STRIPS part of the Planning Domain Definition Language (PDDL) \cite{mcdermott1998pddl}.
Nevertheless, our empirical evaluation results show that FPV performs very well in domains not restricted to STRIPS.
It is important to note that negated facts in the preconditions and goal(s) must be handled properly for the STRIPS principle to be applied to non-STRIPS domains.
Our current implementation compiles negated facts away and introduces an additional fact for each negated fact in the grounded planning domain.

\begin{definition}[(STRIPS) Planning Problem]
    A Planning Problem is a Tuple $P = \langle F, s_0, A, g \rangle$ where $F$ is a set of facts, $s_0 \subseteq F$ and $g \subseteq F$ are the initial state and the goal and $A$ is a set of actions.
    Each action is defined by its preconditions $Pre(a) \subseteq F$ and its effects $Add(a) \subseteq F$ and $Del(a) \subseteq F$.
    $Add(a)$ and $Del(a)$ describe the effects of an action $a$ in terms of facts that are added and deleted from the current state when the planning agent executes $a$.
    Actions have a non-negative cost $c(a)$.
    All facts $f \in F$ that are true in the current planning state define a state $s \subseteq F$.
    A state $s$ is a goal state if and only if $s \supseteq g$.
    An action $a$ is applicable in a state $s$ if and only if $Pre(a) \subseteq s$.
    Applying an action $a$ in a state $s$ leads to a new state $s' = (s \cup Add(a) \setminus Del(a))$.
\end{definition}
\begin{definition}(Solution to a Planning Problem)
    A solution for a planning problem is a sequence of applicable actions $\pi = (a_i)_{i\in [1,N]}$ that transforms $s_0$ into a goal state.
    The cost of a plan is defined as $c(\pi) = \sum \limits_i c(a_i)$.
    A plan is optimal if the cost of the plan is minimal.
\end{definition}

\paragraph{Ignore Delete Effects Relaxation.}
\label{subsec:deleteRelaxation}
In this paper, we propose to use concepts developed in the context of the \emph{Ignore Delete Effects Relaxation}, which has been popular in classical planning since its introduction by \citet{bonet1997robust}.
In the remainder of the paper, we use the term \textit{relaxed planning state} to refer to a delete relaxed planning state.
As the name indicates, this relaxation ignores all delete effects of the actions defined in a planning domain.
More formally, a delete relaxed STRIPS planning problem is defined as follows:
\begin{definition}[Delete Relaxed (STRIPS) Planning Problem]
\label{def:deleteRelaxedPlanningProblem}
    Given a planning problem $P = \langle F, s_0, A, g \rangle$, the corresponding delete relaxed planning problem is defined as $P^+ = \langle F, s_0, A^+, g\rangle$.
    The delete relaxed action set $A^+$ contains all delete relaxed actions from $A$.
    The preconditions, add list, and delete list for the delete relaxed version $a^+$ of action $a$ are defined as $pre_{a^+}=pre_a$, $add_{a^+} = add_a$, $del_{a^+} = \emptyset$.
    Hence, applying an action $a^+$ in a state $s$ leads to a new state $s^+ = (s \cup Add(a^+))$.
\end{definition}
\begin{definition}(Solution to a Delete Relaxed Planning Problem)
    A solution for a delete relaxed planning problem is a sequence of applicable actions (relaxed plan) $\pi^+ = (a_i^+)_{i\in [1,N]}$ that transforms $s_0$ into a goal state.
    The cost of a relaxed plan is defined as $c(\pi^+) = \sum \limits_i c(a_i^+)$.
    A relaxed plan is optimal if the cost of the relaxed plan is minimal.
\end{definition} 

\subsection{Goal Recognition}
\label{subsec:goalRecognition}
This paper investigates a solution method for the (online) goal recognition problem.
Let us first define the goal recognition problem:
\begin{definition}[Goal Recognition]
\label{def:goalRecognition}
    \textit{Goal recognition} is the problem of inferring a nonempty subset $\widehat{G}$ of a set of intended goals $G$ of an observed agent, given a possibly incomplete sequence of observed actions $O$ and a domain model $D$ that describes the environment in which the observed agent acts.
    The observation sequence $O$ is a plan from $s_0$ to the agent's hidden true goal $g^*$.
    Further, the observed agent acts according to a hidden policy $\delta$.
    More formally, a goal recognition problem is a tuple $R = \langle D, O, G \rangle$.
\end{definition}
\begin{definition}(Solution to a Goal Recognition Problem)
    A solution to a goal recognition problem $R$ is a nonempty subset $\widehat{G} \subseteq G$ such that all $g \in \widehat{G}$ are considered to be equally most likely to be the true hidden goal $g_*$ that the observed agent currently tries to achieve.
\end{definition}
The most favorable solution to a goal recognition problem $R$ is a subset $\widehat{G}$ containing only the true hidden goal $g_*$.
In this paper, $D = \langle F, s_0, A \rangle$ is a planning domain with a set of facts $F$, the initial state $s_0$, and a set of actions $A$.
The online goal recognition problem is an extension to the previously defined goal recognition problem that additionally introduces the concept of time, and we define it as follows:
\begin{definition}[Online Goal Recognition]
\label{def:onlineGoalRecognition}
    \textit{Online goal recognition} is a special variant of the \textit{goal recognition} problem, where we assume that the observation sequence $O$ is revealed incrementally.
    More explicitly, let $t \in [1,T]$ be a time index, where $T = |O|$ and hence, the observation sequence for time index $t$ is defined as $O_t = (o_i)_{i\in [1,t]}$.
    For every value of $t$, a goal recognition problem $R(t)$ can be induced as $R(t) = \langle D, G, O_t \rangle$.
\end{definition}
\begin{definition}(Solution to a Online Goal Recognition Problem)
    A solution to the online goal recognition problem are the nonempty subsets $\widehat{G}_t \subseteq G; \forall t \in [1,T]$.
\end{definition}
It is important to note that, in contrast to many existing methods (i.e., \cite{ramirez2009plan}, \cite{ramirez2010probabilistic}, \cite{masters2017cost}, \cite{vered2016mirroring}, \cite{sohrabi2016revisited}), FPV can deal with observing actions \textit{and} observing states simultaneously.
FPV requires fact observations, as FPV compares the observed facts with the fact observation probabilities.
When the observation sequence contains actions, the observed facts can be derived from the given planning domain, which defines the add effects of each action.
In the case of state observations, the observed states directly define the observed facts.

\section{Fact Probability Vector Based Online Goal Recognition}
\label{sec:relaxedSimilarityRecognition}
In this paper, we propose to perform goal recognition by comparing the fact observation probability and the set of actually observed planning facts.
To discuss how the fact observation probability is defined more formally, we first have to define when a planning fact is considered to be observed.
\begin{definition}[Observed Planning Fact]
\label{def:observedFact}
Given a goal recognition problem $R$, we define a planning fact $f\in F$ to be observed during an observation sequence $O$ if and only if $f\in \bigcup_{a\in O}add(a)$.
\end{definition}
\begin{definition}[Fact Observation Probability]
\label{def:expectedFactProbability}
We define the set of fact observation probabilities for each goal $g\in G$ as $\mathcal{P}_F^g = \{P_f(B|s_0,g) | f\in F\}$, where variable $B$ can take two values; one representing that $f$ is observed (cf., Definition \ref{def:observedFact}) and the other representing that $f$ is not observed.
\end{definition}
For example, the distribution $P_f(B|s_0, g)$ models the probability that fact $f$ is observed during an observation sequence that starts in $s_0$ and ends in a goal state $s_g$.

FPV implements the comparison between fact observation probabilities and observed facts based on real vector interpretations of $\mathcal{P}_F^g$, $s_0$, and the currently observed state $s_t$.

\paragraph{Mapping Planning States And $\mathcal{P}_F^g$ Into a Real Vector Space.}
To map a planning state $s$ into its $|F|$-dimensional real vector representation $\mathbf{s} \in \mathbb{R}^{|F|}$, we use the following mapping:
\begin{equation}
\label{eq:RelaxedVectorMapping}
    s_f = \begin{cases}
    1, & iff \hspace{0.2cm} f \in s \\
    0, & else
    \end{cases}
\end{equation}
In Equation \ref{eq:RelaxedVectorMapping}, $s$ is a planning state, and $\mathbf{s}_f$ is the element of $\mathbf{s}$ that encodes the planning fact $f$.
The resulting vector mapping $\mathbf{s}$ of a planning state $s$ encodes the observational evidence from $s$ in the frame of fact observation probabilities.
Algorithm \ref{alg:FPVGR}, presented later in the paper, will refer to this function under the signature $mapState(s)$.

To map the fact observation probabilities $\mathcal{P}_F^g$ (cf., Definition \ref{def:expectedFactProbability} into an $|F|$ dimensional real vector representation $\mathbf{v}^{g} \in\mathbb{R}^{|F|}$, given an initial state $s_0$ and a goal description $g$, FPV uses the following mapping:
\begin{equation}
\label{eq:ProbabilityMapping}
    \mathbf{v}^{g}_{f} = P_f(f\in \bigcup_{a\in O}add(a)|s_0, g)
\end{equation}
Algorithm \ref{alg:FPVGR}, which is presented later in the paper, will refer to this function under the signature $mapP(s_0, g, \mathcal{P}_F^g)$.
As an example, consider a simple domain in which $F = \{f_1, f_2, f_3\}$, $s_0 = \emptyset$, and there is only one possible goal $g = \{f_3\}$.
Further, $\mathcal{P}_F^g$ is defined as $\mathcal{P}_F^g=\{P_{f_1}=(0.8,0.2),P_{f_2}=(0.3, 0.7),P_{f_3}=(0.6, 0.4)\}$, where we use the following notation $(P_f(f\in \bigcup_{a\in O}add(a)|s_0, g), P_f(f\not\in \bigcup_{a\in O}add(a)|s_0, g))$.
For this example, $\mathbf{v}^g$ is defined as $\mathbf{v}^g = (0.8,0.3,0.6)$.
The resulting vector representations of $\mathcal{P}_F^g$ can be interpreted as an encoding of how probable it is that each fact $f\in F$ is added to the current planning state $s_t$ at any point in time by an observation sequence that starts at $s_0$ and leads to goal $g$.

\paragraph{Method.}
For goal recognition, FPV calculates a heuristic score for each goal based on the mappings of the points $\mathbf{s_0^+}$ and $\mathbf{s_t^+}$ (cf., Equation \ref{eq:RelaxedVectorMapping}) and the mapping of $\mathcal{P}_F^g$ for each goal (cf., Equation \ref{eq:ProbabilityMapping}).
FPV uses the currently observed relaxed planning state (i.e., $s_t^+$) instead of the non-relaxed state because it contains additional information about the planning states the observation sequence has already visited.
This is because, as defined by Definition \ref{def:deleteRelaxedPlanningProblem}, under the delete relaxation, facts can only be added to the initial planning state but can never be deleted.
Consequently, $s_t^+$ contains all facts added by any action in the relaxed observation sequence $O^+$, which contains the relaxed action $a^+$ for all actions $a\in O$.
The resulting vector $\mathbf{s_t^+}$ contains a one for all facts that \textit{either} have been already true in $s_0$ \textit{or} were added by an action in the observation sequence and zero otherwise.
Based on the vector mappings, FPV calculates the goal recognition heuristic score for each goal as defined by Equation \ref{eq:heuristicCalculation}.
\begin{equation}
\label{eq:heuristicCalculation}
    h(\mathbf{s_0^+}, \mathbf{s_t^+}, \mathbf{v}^{g}) = ||\overrightarrow{(\mathbf{s_0^+}\odot \mathbf{v}^{g},\mathbf{v}^{g})}||_2 - ||\overrightarrow{(\mathbf{s_t^+}\odot \mathbf{v}^{g},\mathbf{v}^{g})}||_2
\end{equation}
In Equation \ref{eq:heuristicCalculation}, $\odot$ is a special elementwise multiplication of two vectors as defined by Equation \ref{eq:elementWiseMultiplication}, $\overrightarrow{(\mathbf{x},\mathbf{y})}$ is a direction vector between two points $\mathbf{x}$ and $\mathbf{y}$ as defined by Equation \ref{eq:directionVector}, and $||\mathbf{x}||_2$ is the $l^2$ vector norm, which is defined as $||\mathbf{x}||_2 = \sqrt{\sum_{k=1}^n{x_k^2}}$.
\begin{equation}
\label{eq:elementWiseMultiplication}
    (\mathbf{s} \odot \mathbf{v})^i = \begin{cases}
    \mathbf{s}^i * \mathbf{v}^i, & iff \hspace{.2cm} \mathbf{v}^i > 0 \\
    \mathbf{s}^i, & else
    \end{cases}    
\end{equation}
Using the elementwise multiplication $\odot$, as defined by Equation \ref{eq:elementWiseMultiplication}, ensures that the heuristic value monotonically increases for a goal $g$ when only facts for which $P_f(f\in \bigcup_{a\in O}add(a)|s_0, g) > 0$ holds are observed.
\begin{equation}
\label{eq:directionVector}
    \overrightarrow{(\mathbf{x}, \mathbf{y})} = \mathbf{y} - \mathbf{x}
\end{equation}
Computing the direction between $(\mathbf{s_0^+}\odot\mathbf{v}^g)$ and $\mathbf{v}^g$ results in a vector that encodes what has to be changed from the initial state to reach a valid goal state as approximated by $\mathcal{P}_F^g$.
This is because the resulting vector becomes zero at all points where \textit{both} $\mathbf{s_0^+}$ and $\mathbf{v}^g$ have values \textit{greater than} zero.
Similarly to the first direction vector, computing the direction between $(\mathbf{s_t^+}\odot \mathbf{v}^{g})$ and $\mathbf{v}^g$ results in a vector that encodes what has to be changed from the currently observed relaxed state $\mathbf{s_t^+}$ to reach a valid goal state.
Hence, the heuristic computed by FPV estimates the probability of a goal $g$ being the true hidden goal $g^*$ of the observed agent by computing the distance already covered by the observation sequence between the initial state and a valid goal state for $g$ as estimated by the fact observation probabilities.
In addition, the heuristic punishes goals for which $\mathcal{P}_F^g$ assigns a \textit{probability of zero} to facts that were \textit{actually observed}.
This is because, in this case, the vector resulting from $\overrightarrow{(\mathbf{s_t^+}\odot \mathbf{v}^{g},\mathbf{v}^{g})}$ will have a value of minus one at all positions that correspond to such facts.
Consequently, $||\overrightarrow{(\mathbf{s_t^+}\odot \mathbf{v}^{g},\mathbf{v}^{g})}||_2$ increases, which results in an overall decreased heuristic value.

\begin{algorithm}[t!]
\caption{Fact Observation Probability Based Goal Recognition.}
\label{alg:FPVGR}
\begin{algorithmic}[1]
\Function{Recognize goals}{$\mathcal{P}_F$, $s_0$, $G$, $O_t$}
\State $\widehat{G} \leftarrow \emptyset$ \Comment{Set of recognized goals}
\State $h \leftarrow \{\}$ \Comment{Maps goals to heuristic values}
\ForAll{$g \in G$}
    \State $\mathbf{v}^{g} \leftarrow mapP(s_0, g, \mathcal{P}_F^g)$ \Comment{cf., Equation \ref{eq:ProbabilityMapping}}
\EndFor
\State $s_t^+ \leftarrow s_0^+$ \Comment{Initialize current observed state}
\ForAll{$o_i\in O_t$}
    \State $s_t^+ \leftarrow s_t^+[[o_i]]$ \Comment{Update observed state}
\EndFor
\State $\mathbf{s_t^+} \leftarrow mapState(s_t^+)$ \Comment{cf., Equation \ref{eq:RelaxedVectorMapping}}
\ForAll{$g \in G$}
    \State $h[g] \leftarrow h(\mathbf{s_0^+}, \mathbf{s_t^+}, \mathbf{v}^{g})$ \Comment{cf., Equation \ref{eq:heuristicCalculation}}
\EndFor
\State $maxH \leftarrow maxValue(h)$
\ForAll{$g \in G$}
    \If{$h[g] = maxH$}
        \State $\widehat{G} \leftarrow \widehat{G} \cup \{g\}$
    \EndIf
\EndFor
\State \Return $\widehat{G}$
\EndFunction
\end{algorithmic}
\end{algorithm}
The algorithm used to recognize goals based on the previously defined vector mappings is summarized in Algorithm \ref{alg:FPVGR}.
As a first step, $\mathbf{v}^{g}$ has to be determined for each goal $g \in G$ (lines 4-6).
As a second step, the vector representation $\mathbf{s_t^+}$ of the currently observed relaxed state is calculated (lines 7-9).
Afterward, Algorithm \ref{alg:FPVGR} calculates the heuristic value for all goals (lines 10-12).
As a last step, from the calculated heuristic values, Algorithm \ref{alg:FPVGR} selects the goals that have the maximum value (lines 13-19):

\begin{figure}[h]
\centering
\includegraphics[width=0.4\textwidth]{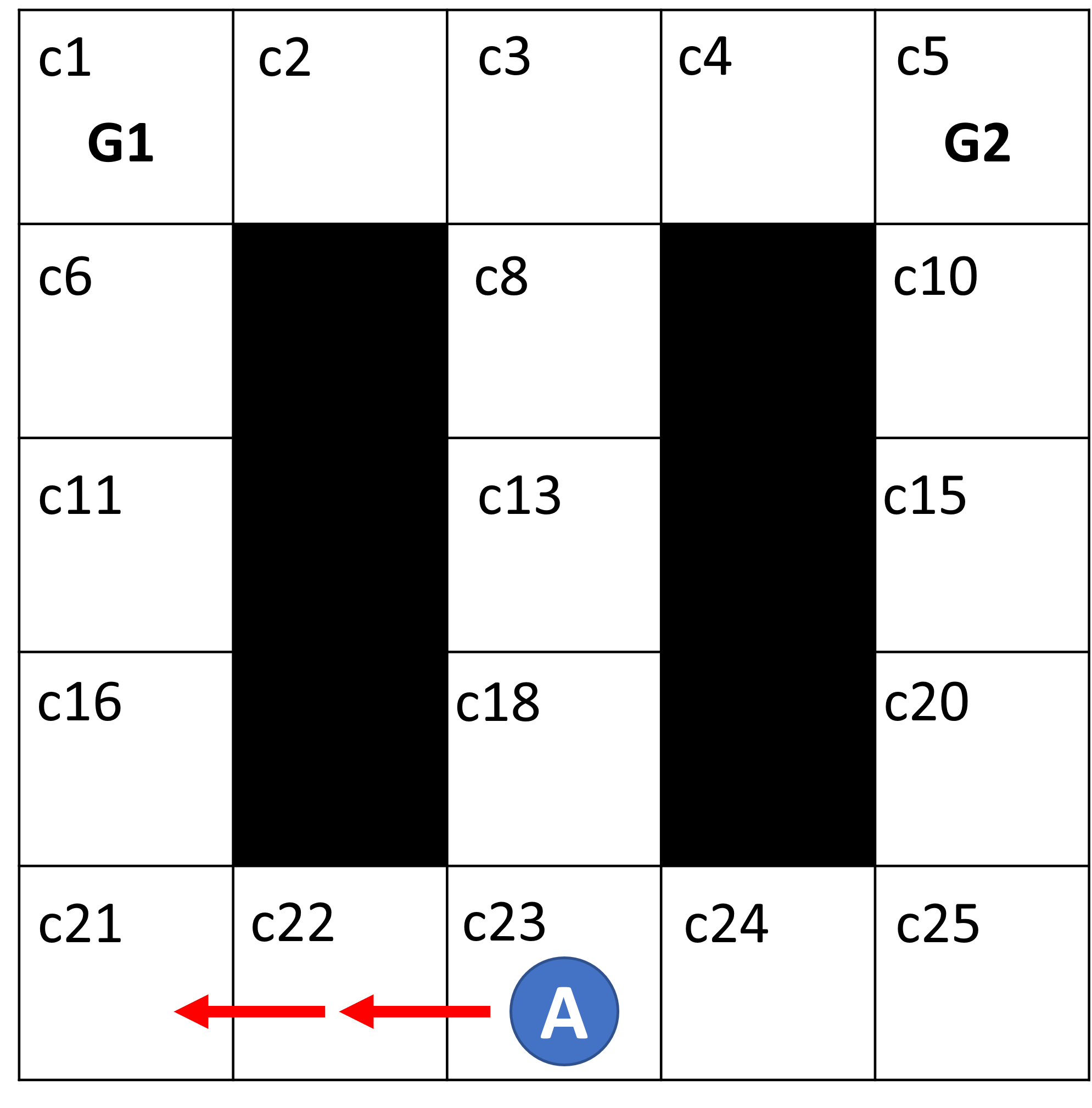}
\caption{Exemplary grid environment. \vspace{.7cm}}
\label{fig:RunningExample}
\end{figure}

\paragraph{Example.}
This paragraph illustrates FPV based on a simple example.
For this example, we consider the grid environment in Figure \ref{fig:RunningExample}.
In \ref{fig:RunningExample}, the agent is initially located in cell c23 and wants to reach one of the two goals, $G1$ or $G2$.
The agent can move through the environment using moves in all directions \emph{except diagonal moves} to all cells that are not colored black; each move has a cost of 1.
We assume that Figure \ref{fig:RunningExample} visualizes a corresponding planning domain model.
This domain model includes only one predicate (is-at ?x) that models the current position of the agent (e.g., (is-at c23)) and only one action schema m(?x, ?y) that allows the agent to move from cell x to cell y.
Further, the red arrows indicate actions that have already been observed.
In addition, we assume that the agent acts entirely rationally according to the modeled action costs (i.e., the agent uses cost-optimal plans).

In this example, both goals have exactly two optimal plans.
Further, we assume that the agent is indifferent about taking any of the two optimal plans and, hence, takes either path with a probability of 0.5.
Table \ref{tab:runningExampleProbabilities} summarizes the fact observation probabilities for the described example domain.
Consequently, for this example, $\mathbf{v}^{G1}$ equals the observed column for $P_f^{G1}$ from top to bottom and $\mathbf{v}^{G2}$ the observed column for $P_f^{G2}$ from top to bottom.
The two red arrows in Figure \ref{fig:RunningExample} define the observation sequence for this scenario as $O=\{m(c23,c22), m(c22,c21)\}$.
Hence, the currently observed relaxed state is $s_t^+ = \{$(is-at c23), (is-at c22), (is-at c21)$\}$.
For goal recognition, FPV calculates the heuristic values for G1 and G2.
We will exemplify the calculation for G1 only here, as the calculation follows similar paths for G2.
As a first step $||\overrightarrow{(\mathbf{s_0^+}\odot\mathbf{v}^{G1},\mathbf{v}^{G1})}||_2$ is calculated.
In this example, as only one fact (i.e., (is-at c23)) is true in $s_0$, $\mathbf{s_0^+}$ has only one non-zero element representing the fact (is-at c23).
Hence, the result of $\mathbf{s_0^+}\odot\mathbf{v}^{G1}$ is similar to $\mathbf{s_0^+}$ as the fact observation probability for (is-at c23) is equal to 1 and all other elements of $\mathbf{v}^{G1}$ are multiplied out.
From this, FPV calculates $\overrightarrow{(\mathbf{s_0^+}\odot\mathbf{v}^{G1},\mathbf{v}^{G1})}$.
The resulting vector is, in this example, equal to $\mathbf{v}^{G1}$ except for having a value of 0 at the position that represents the fact (is-at c23) as this has a value of 1 in both vectors.
Then, FPV calculates the $l^2$ norm for this vector, which is 1.87 for this example.
As a second step $||\overrightarrow{(\mathbf{s_t^+}\odot\mathbf{v}^{G1},\mathbf{v}^{G1})}||_2$ is calculated.
In contrast to $\mathbf{s_0^+}$, $\mathbf{s_t^+}$ has three non-zero elements representing the facts in $s_t^+$ (i.e., (is-at c23), (is-at c22), and (is-at c21)).
Consequently, the result of $\mathbf{s_t^+}\odot\mathbf{v}^{G1}$ also has three non-zero elements containing the observation probabilities for the facts (is-at c23), (is-at c22), and (is-at c21).
Hence, $\overrightarrow{(\mathbf{s_t^+}\odot\mathbf{v}^{G1},\mathbf{v}^{G1})}$ is equal to $\mathbf{v}^{G1}$ except for having a value of 0 for the three facts in $s_t^+$.
As a result, this vector's norm is 1.73, which leads to a heuristic value $h(\mathbf{s_0^+}, \mathbf{s_t^+}, \mathbf{v}^g) = 1.87 - 1.73 = 0.14$.
As the observation sequence does not add facts assigned a value larger than 0 by the fact observation probabilities for G2, both parts of the heuristic computation for G2 will result in the same vector norm.
Hence, for G2 we have a heuristic value of 0 in this example.
Consequently, FPV recognizes G1 as the most probable goal in this example.

\begin{table}[t!]
\caption{Fact observation probabilities for the example depicted in Figure \ref{fig:RunningExample}.}
\label{tab:runningExampleProbabilities}
\centering
\begin{tabular}{l|ll|ll}
    $f$        & $P_f^{G1}$       &              & $P_f^{G2}$       &              \\
            & observed & not observed & observed & not observed \\ \hline\hline
(is-at c1)  & 1.0      & 0.0          & 0.0      & 1.0          \\
(is-at c2)  & 0.5      & 0.5          & 0.0      & 1.0          \\
(is-at c3)  & 0.5      & 0.5          & 0.5      & 0.5          \\
(is-at c4)  & 0.0      & 1.0          & 0.5      & 0.5          \\
(is-at c5)  & 0.0      & 1.0          & 1.0      & 0.0          \\
(is-at c6)  & 0.5      & 0.5          & 0.0      & 1.0          \\
(is-at c7)  & 0.0      & 1.0          & 0.0      & 1.0          \\
(is-at c8)  & 0.5      & 0.5          & 0.5      & 0.5          \\
(is-at c9)  & 0.0      & 1.0          & 0.0      & 1.0          \\
(is-at c10) & 0.0      & 1.0          & 0.5      & 0.5          \\
(is-at c11) & 0.5      & 0.5          & 0.0      & 1.0          \\
(is-at c12) & 0.0      & 1.0          & 0.0      & 1.0          \\
(is-at c13) & 0.5      & 0.5          & 0.5      & 0.5          \\
(is-at c14) & 0.0      & 1.0          & 0.0      & 1.0          \\
(is-at c15) & 0.0      & 1.0          & 0.5      & 0.5          \\
(is-at c16) & 0.5      & 0.5          & 0.0      & 1.0          \\
(is-at c17) & 0.0      & 1.0          & 0.0      & 1.0          \\
(is-at c18) & 0.5      & 0.5          & 0.5      & 0.5          \\
(is-at c19) & 0.0      & 1.0          & 0.0      & 1.0          \\
(is-at c20) & 0.0      & 1.0          & 0.5      & 0.5          \\
(is-at c21) & 0.5      & 0.5          & 0.0      & 1.0          \\
(is-at c22) & 0.5      & 0.5          & 0.0      & 1.0          \\
(is-at c23) & 1.0      & 0.0          & 1.0      & 0.0          \\
(is-at c24) & 0.0      & 1.0          & 0.5      & 0.5          \\
(is-at c25) & 0.0      & 1.0          & 0.5      & 0.5          
\end{tabular}
\end{table}

\section{Estimating Fact Observation Probabilities}
\label{sec:estimation}
In practice, the fact observation probabilities are usually not given.
Hence, this section proposes a method to estimate the fact observation probabilities from the given planning domain.
To estimate fact observation probabilities, FPV first determines sets of supporter actions for each possible goal.
Supporter actions were introduced by \citet{hoffmann2004ordered}, and they can be sampled very efficiently from a Relaxed Planning Graph (RPG)\footnote{An RPG is created by sequentially adding all applicable actions to the initial state in a relaxed way until all goal facts are relaxed reachable.} \cite{hoffmann2004ordered}.
The idea behind supporter actions was to define a methodology for determining the actions in a given planning domain that are relevant to solving a given planning problem.
More precisely, an action $a$ is called a supporter for a fact $f$ if $f\in add(a)$.
The following subsection describes the exact algorithm that FPV uses to sample supporter actions.
In total, FPV generates $N$ different sets of supporter actions for each goal to ensure that the sets of supporter actions cover several possible paths.
Based on the $N$ sets of supporter actions for each goal, FPV calculates for each action $a$ how probable it is that $a$ occurs in one of the $N$ sets.
These probabilities can be interpreted as action observation probabilities from which the fact observation probabilities can be derived.
Given these probabilities, FPV derives the observation probability for each $f\in F$ (i.e., $\mathbf{P}_F^g$) by calculating the probability of observing \textit{any} of the actions for which $f\in add(a)$ holds.

\subsection{Sampling Supporter Actions}
To generate the $N$ sets of supporter actions for a goal description $g$, FPV first generates $N$ sets of supporter actions for each subgoal $g_i\in g$.
Algorithm \ref{alg:findSupporters} describes the exact algorithm that FPV uses to generate the $N$ sets of supporters for each subgoal $g_i$.
\begin{algorithm}[t!]
\caption{Supporter Action Sampling.}
\label{alg:findSupporters}
\begin{algorithmic}[1]
\Function{SampleRelevantActions}{$g_i$, $RPG$, $s_0$, $N$}
\State $count \leftarrow \{\}$ \Comment{Map from action to count in samples.}
\State $samples \leftarrow []$ \Comment{List of generated supporter sets.}
\ForAll{$i \in range(0,N)$}
    \State $C \leftarrow g_i$ \Comment{Facts to be supported.}
    \State $found \leftarrow \emptyset$ \Comment{Stores supported facts.}
    \State $sups \leftarrow \emptyset$ \Comment{Stores supporters.}

    \For{$t =$ $RPG.levels$ to $0$}
        \State $newC \leftarrow \emptyset$ \Comment{New facts to be supported.}
        \While{$|C| > 0$}
            \State $p \leftarrow C.pop()$
            \State $psups \leftarrow \emptyset$ \Comment{Potential supporters.}
            \For{$t2$ = $0$ to $t$}
                \ForAll{$a \in RPG.level(t2)$}
                    \If{$|add(a)\cap p| > 0$}
                        \State $psups \leftarrow psups \cup \{a\}$
                    \EndIf
                \EndFor
                \If{$|psups| > 0$}
                    \State break
                \EndIf
            \EndFor
            \State $psups \leftarrow minCount(psups, count)$
            \State $a \leftarrow random(psups)$
            \State $found \leftarrow found \cup \{p\}$
            \State $C \leftarrow C \setminus \{p\}$
            \State $sups \cup \{a\}$
            \State $count[a] \leftarrow count[a] + 1$
            \ForAll{$n \in pre(a)$}
                \If{$n \not\in s_0 \land n \not\in found \land n \not\in C$}
                    \State $newC \leftarrow newC \cup \{n\}$
                \EndIf
            \EndFor
            \ForAll{$r \in add(a)$}
                \State $C \leftarrow C \setminus \{r\}$
                \State $newC \leftarrow newC \setminus \{r\}$
            \EndFor
        \EndWhile
        \State $C \leftarrow C \cup \{newC\}$
    \EndFor
    \State $samples \leftarrow samples.add(sups)$
\EndFor
\State \Return $samples$
\EndFunction
\end{algorithmic}
\end{algorithm}
The input of Algorithm \ref{alg:findSupporters} consists of four elements: $f$ is the planning fact for which the sets of supporter actions are sampled, $RPG$ is an RPG for the given planning problem, $s_0$ is the initial state, and $N$ is the number of supporter action sets that are generated.
In lines 2 and 3, the $count$ map and $samples$ list are initialized.
The $count$ map has actions as keys and maps each action key to the number of times that action has already been selected as a supporter action.
The $samples$ is the final return variable of the algorithm and stores the generated sets of supporter actions.
The for loop that starts in line 4 repeats the following sampling procedure $N$ times.
For the sampling process, first the elements $C$, $found$, and $sups$ are initialized (lines 5-7).
$C$ is the set of facts for which the algorithm samples supporting actions, $found$ is a set that keeps track of the facts already supported, and $sups$ is the set of supporting actions already selected for the current sample.
The search for supporting actions for a fact $p$ occurs in lines 13-22.
For this search, the algorithm starts at the first action level of the RPG, which ensures that supporters closer to $s_0$ are found first.
All potential supporter actions for $p$ from the currently searched RPG action level are added to the $psups$ set.
When $psups$ contains at least one element after searching the current RPG action level, no further levels are searched.
In Line 23, the algorithm selects the potential supporter action that has been selected the minimum number of times (i.e., it has the minimal count value compared to all elements of $psups$).
This selection procedure ensures that when different options exist for selecting supporting actions, all options are picked with roughly the same probability.
When there are several actions with the minimum count value, the algorithm randomly selects one of those actions (Line 24).
Afterward, the $found$, $C$, $sups$, and $count$ variables are updated accordingly (lines 25-28).
In lines 29-33, the algorithm iterates over the preconditions of the selected action and adds all facts that are not already supported or part of $s_0$ to $newC$.
Following this, in lines 34-37, the algorithm removes all add effects of the selected action from $C$ and $newC$ as the selected action already supports them.
As a result, Algorithm \ref{alg:findSupporters} returns a list of sampled sets of relevant actions for reaching the subgoal $g_i$ from $s_0$.

\begin{algorithm}[h!]
\caption{Generating Supporter Actions for $g$.}
\label{alg:generateSupporterActions}
\begin{algorithmic}[1]
\Function{Recognize goals}{$S$, $N$, $g$}
\State $SG \leftarrow []$ \Comment{List of supporter action sets.}
\State $SG_i \leftarrow \emptyset$ \Comment{Set of supporter actions.}
\ForAll{$x \in range(0, N)$}
    \ForAll{$g_i \in g$}
        \State $s \leftarrow random(S[g_i])$
        \State $SG_i \leftarrow SG_i \cup s$
        \State $S[g_i] \leftarrow S[g_i] \setminus \{s\}$
    \EndFor
    \State $SG \leftarrow SG.add(SG_i)$
    \State $SG_i \leftarrow \emptyset$
\EndFor
\State \Return $SG$
\EndFunction
\end{algorithmic}
\end{algorithm}
From these sets of supporter actions for each subgoal, FPV then generates $N$ sets of supporter actions for $g$.
FPV does this by using Algorithm \ref{alg:generateSupporterActions}.
The input of Algorithm \ref{alg:generateSupporterActions} is composed of three elements:
The first element, $S$, is a map that maps all subgoals $g_i$ to their respective list of supporter action sets sampled previously by Algorithm \ref{alg:findSupporters}.
The second input element, $N$, is the number of supporter action sets generated.
The third input element, $g$, is the goal description.
In the beginning, Algorithm \ref{alg:generateSupporterActions} initializes the two sets $SG$ and $SG_i$ in lines 2 and 3.
$SG$ stores the generated sets of supporter actions for $g$, and $SG_i$ is used as a temporary set to unify the sets of supporter actions from each subgoal $g_i$.
In lines 4 to 12, the algorithm iterates $N$ times to generate $N$ sets of supporter actions from $g$.
This is done by randomly picking one set of supporter actions for each subgoal $g_i$ from $S$ (Line 6) and unifying it with $SG_i$ (Line 7).
Removing all already-picked sets of supporter actions from $S$ ensures that each set is only picked once (Line 8).
After the for loop, which ends in Line 9, $SG_i$ holds a complete supporter action set for $g$ that the algorithm then adds to $SG$.
As a result, Algorithm \ref{alg:generateSupporterActions} returns a set of $N$ supporter action sets for $g$.

\section{Evaluation}
\label{sec:evaluation}
We empirically evaluated the FPV method on 15 benchmark domains commonly used to evaluate goal recognition approaches.
The empirical evaluation aims to achieve the following goals:
\begin{enumerate}
    \item Evaluate the goal recognition performance of the FPV method compared to four existing goal recognition methods.
    \item Investigate how efficient the FPV method is regarding computation time compared to four existing goal recognition methods.
\end{enumerate}
All experiments consider the \emph{online} goal recognition problem, as defined by Definition \ref{def:onlineGoalRecognition}.

\paragraph{Datasets.}
We evaluated the approaches on 15 commonly used benchmark domains (e.g., \citet{pereira2020landmark}).

\paragraph{Goal Recognition Methods.}
To compare the performance of the FPV method to the performance of existing goal recognition methods, we implemented a planning-landmark-based approach introduced by \citet{pereira2020landmark} (PLR), an approximate approach proposed by \citet{ramirez2009plan} (RG09), an LP Based approach introduced by \citet{santos2021lp} (LP), and an approach by \citet{masters2017cost} (MS).
FPV, PLR, and MS were implemented in Java using the PPMAJAL\footnote{\url{https://gitlab.com/enricos83/PPMAJAL-Expressive-PDDL-Java-Library}} library as a basis for PDDL related functionalities.
For the RG09 and LP approaches, we used the original code that is publicly available. 
We chose the PLR, LP, and MS approaches because they are recent methods for the two major types of methods that implement the PRAP principle, i.e., approaches that derive a probability distribution over the possible goals by a comparison among the costs of different plans that are calculated for each goal (MS) and approaches that use heuristic scores to rank all possible goals $g \in G$ (PLR and LP).
In addition, we compare FPV to the RG09 approach because it is related to the FPV method through the use of concepts from relaxed planning.
For the MS approach, we used the MetricFF planner \cite{hoffmann2003metric} with $\beta=1$.
For the FPV method, we sample $N=10$ relevant action sets for each $g$ and, as FPV involves some random selections, we report the results averaged over 20 repeated runs.
We also tested larger values for $N$, which did not significantly increase recognition performance.

\paragraph{Experimental Design.}
All experiments in this evaluation were carried out on machines with 24 cores, 2.60GHz, and at least 386GB RAM.
We used the mean goal recognition precision to assess the performance of the different methods.
We calculate the precision similar to existing works (e.g., \cite{amado2023robust}).
Furthermore, as we consider online goal recognition problems $R(t)$ in this evaluation, we calculated the mean precision for different fractions $\lambda = t/T$ of total observations used for goal recognition.
Here, we used relative numbers because the lengths of the involved observation sequences substantially differ.
Hence, the mean precision $Prec$ for a fraction $\lambda \in [0, 1]$ is calculated as follows:
\begin{equation}
\label{eq:Precision}
    Prec(\lambda,\mathcal{D}) = \frac{\sum_{R \in \mathcal{D}}{\frac{[g_{*R} \in R(\lfloor T_{R}\lambda\rfloor)]}{|\mathbf{G}_{R(T_R \lambda)}|}}}{|\mathcal{D}|}
\end{equation}
Here, $\mathcal{D}$ is a set of online goal recognition problems $R$, $g_{*R}$ denotes the correct goal of goal recognition problem $R$, $T_R$ is the maximum value of $t$ for online goal recognition problem $R$ (i.e., length of observation sequence that is associated with $R$), and $[g_{*R} \in R(t)]$ equals one if $g_{*R} \in \mathbf{G}_{R(t)}$ and 0 otherwise, where $\mathbf{G}_{R(t)}$ is the set of recognized goals for $R(t)$.
In other words, the precision quantifies the probability of picking the correct goal from a predicted set of goals $\mathbf{G}$ by chance.

\begin{table}[t!]
\caption{Comparison of evaluation results among all evaluated methods when applied to different benchmark domains. The precision is reported for different $\lambda$ values. Column \textbf{S} reports the average spread (size of the set of goals that are recognized as the true goal) for each domain.}

\label{tab:evaluationTable}
\resizebox*{\columnwidth}{!}{
\begin{tabular}{ll|llllllllll|l}
\multicolumn{1}{c}{}		 & \multicolumn{1}{c|}{}		 & \multicolumn{10}{c|}{\textbf{Precision (for different $\lambda$)}}		 & \textbf{S} \\
			 & 			 & .1 		 & .2 		 & .3 		 & .4 		 & .5 		 & .6 		 & .7 		 & .8 		 & .9 		 & 1 		 & \\ \hline 
\multirow{5}{*}{\rotatebox[origin=c]{90}{\textbf{average}}}
 & MS & .35 & .43 & .49 & .57 & .63 & .67 & .73 & .77 & .79 & .81 & 1.9\\
 & RG09 & .32 & .43 & .5 & .59 & .65 & .69 & .75 & .78 & .82 & .86 & 2.2\\
 & LP & .23 & .34 & .42 & .51 & .59 & .64 & .7 & .74 & .81 & .86 & 2.6\\
 & PLR & .3 & .35 & .43 & .51 & .59 & .66 & .7 & .76 & .83 & .9 & 1.2\\
 & FPV(ours) & \textbf{.38} & \textbf{.49} & \textbf{.59} & \textbf{.66} & \textbf{.72} & \textbf{.77} & \textbf{.83} & \textbf{.87} & \textbf{.91} & \textbf{.94} & \textbf{1.1} \\
\hline
\hline
\multirow{5}{*}{\rotatebox[origin=c]{90}{blocks-world}}
 & MS & .1 & .13 & .14 & .24 & .29 & .36 & .41 & .47 & .54 & .55 & 7.1\\
 & RG09 & .08 & .22 & .23 & .29 & .38 & .42 & .53 & .63 & \textbf{.81} & \textbf{.9} & 4.1\\
 & LP & .06 & .13 & .15 & .2 & .33 & .38 & .46 & .52 & .76 & .81 & 5.4\\
 & PLR & .09 & .09 & .1 & .1 & .09 & .09 & .09 & .07 & .21 & .27 & 2.8\\
 & FPV(ours) & \textbf{.13} & \textbf{.31} & \textbf{.24} & \textbf{.32} & \textbf{.42} & \textbf{.47} & \textbf{.57} & \textbf{.66} & .8 & \textbf{.9} & \textbf{1.5} \\
\hline
\multirow{5}{*}{\rotatebox[origin=c]{90}{campus}}
 & MS & .73 & .73 & .73 & .53 & .57 & .43 & .57 & .63 & .6 & .6 & 1.4\\
 & RG09 & .73 & .8 & .8 & .83 & .83 & .83 & .83 & .8 & .8 & .8 & 1.2\\
 & LP & .5 & .87 & .87 & .97 & .97 & .97 & .97 & .97 & .97 & .97 & 1.2\\
 & PLR & \textbf{1.0} & \textbf{1.0} & \textbf{1.0} & \textbf{1.0} & \textbf{1.0} & \textbf{1.0} & \textbf{1.0} & \textbf{1.0} & \textbf{1.0} & \textbf{1.0} & \textbf{1.0} \\
 & FPV(ours) & .67 & .87 & .87 & \textbf{1.0} & \textbf{1.0} & \textbf{1.0} & \textbf{1.0} & \textbf{1.0} & \textbf{1.0} & \textbf{1.0} & \textbf{1.0} \\
\hline
\multirow{5}{*}{\rotatebox[origin=c]{90}{depots}}
 & MS & \textbf{.21} & .16 & .16 & .22 & .38 & .49 & .48 & .54 & .59 & .59 & 1.8\\
 & RG09 & .16 & \textbf{.29} & \textbf{.28} & \textbf{.32} & .45 & .62 & .59 & .6 & .61 & .78 & 2.6\\
 & LP & .09 & .16 & .17 & .27 & .34 & .44 & .57 & .69 & .75 & .89 & 2.4\\
 & PLR & .18 & .14 & .2 & .2 & .23 & .3 & .5 & .64 & .61 & .82 & 1.2\\
 & FPV(ours) & .19 & .18 & \textbf{.28} & .3 & \textbf{.55} & \textbf{.68} & \textbf{.84} & \textbf{.9} & \textbf{.99} & \textbf{1.0} & \textbf{1.0} \\
\hline
\multirow{5}{*}{\rotatebox[origin=c]{90}{driverlog}}
 & MS & .29 & \textbf{.37} & .5 & \textbf{.63} & \textbf{.73} & \textbf{.84} & \textbf{.92} & \textbf{.91} & \textbf{.91} & \textbf{.96} & 1.4\\
 & RG09 & .22 & .27 & .35 & .46 & .52 & .6 & .75 & .77 & .81 & .81 & 2.9\\
 & LP & .2 & .26 & .37 & .5 & .58 & .68 & .67 & .68 & .76 & .8 & 2.3\\
 & PLR & .25 & .23 & .39 & .52 & .61 & .75 & .8 & .8 & .82 & .92 & 1.1\\
 & FPV(ours) & \textbf{.32} & .35 & \textbf{.58} & .61 & .56 & .63 & .79 & .86 & .88 & .88 & \textbf{1.0} \\
\hline
\multirow{5}{*}{\rotatebox[origin=c]{90}{dwr}}
 & MS & .13 & .23 & .2 & .34 & .34 & .39 & .43 & .47 & .46 & .55 & 2.0\\
 & RG09 & .25 & .33 & .32 & .26 & .2 & .22 & .3 & .27 & .31 & .41 & 2.7\\
 & LP & .19 & .18 & .24 & .29 & .28 & .33 & .51 & .51 & .65 & .81 & 2.0\\
 & PLR & .23 & .16 & .27 & \textbf{.45} & \textbf{.5} & \textbf{.54} & .52 & \textbf{.63} & \textbf{.79} & \textbf{1.0} & 1.1\\
 & FPV(ours) & \textbf{.31} & \textbf{.49} & \textbf{.47} & .44 & .36 & .44 & \textbf{.61} & .61 & .73 & .87 & \textbf{1.0} \\
\hline
\multirow{5}{*}{\rotatebox[origin=c]{90}{easy-ipc-grid}}
 & MS & .29 & .32 & .44 & \textbf{.52} & .56 & .63 & \textbf{.71} & .76 & .83 & .94 & 2.4\\
 & RG09 & \textbf{.32} & \textbf{.34} & .45 & .48 & .57 & .63 & .69 & \textbf{.78} & .85 & \textbf{1.0} & 3.0\\
 & LP & .13 & .24 & .28 & .35 & .39 & .47 & .5 & .57 & .7 & .78 & 3.3\\
 & PLR & .2 & .22 & .36 & .42 & .42 & .5 & .58 & .59 & .72 & \textbf{1.0} & \textbf{1.0} \\
 & FPV(ours) & .28 & .33 & \textbf{.46} & \textbf{.52} & \textbf{.67} & \textbf{.67} & .69 & \textbf{.78} & \textbf{.88} & .93 & \textbf{1.0} \\
\hline
\multirow{5}{*}{\rotatebox[origin=c]{90}{ferry}}
 & MS & .38 & \textbf{.7} & \textbf{.8} & .83 & .89 & .95 & .93 & .96 & .96 & .96 & 1.2\\
 & RG09 & .28 & .58 & .73 & .8 & .89 & .91 & .95 & .96 & .96 & .96 & 1.6\\
 & LP & .19 & .34 & .54 & .75 & .85 & .88 & .91 & .91 & .93 & .96 & 2.1\\
 & PLR & .25 & .38 & .64 & .73 & .84 & .89 & .96 & \textbf{1.0} & \textbf{1.0} & \textbf{1.0} & \textbf{1.1} \\
 & FPV(ours) & \textbf{.41} & .65 & .75 & \textbf{.87} & \textbf{.98} & \textbf{.98} & \textbf{1.0} & \textbf{1.0} & \textbf{1.0} & \textbf{1.0} & 1.2\\
\hline
\multirow{5}{*}{\rotatebox[origin=c]{90}{intrusion}}
 & MS & \textbf{.26} & \textbf{.45} & .59 & \textbf{.8} & \textbf{.83} & \textbf{.83} & \textbf{1.0} & \textbf{1.0} & \textbf{1.0} & \textbf{1.0} & 1.8\\
 & RG09 & \textbf{.26} & \textbf{.45} & .59 & \textbf{.8} & \textbf{.83} & \textbf{.83} & \textbf{1.0} & \textbf{1.0} & \textbf{1.0} & \textbf{1.0} & 1.8\\
 & LP & .18 & .24 & .35 & .38 & .66 & .67 & .72 & .78 & .94 & .96 & 2.9\\
 & PLR & .16 & .33 & .34 & .62 & .64 & .77 & .81 & .92 & \textbf{1.0} & \textbf{1.0} & \textbf{1.1} \\
 & FPV(ours) & .22 & .4 & \textbf{.65} & .77 & .81 & .81 & \textbf{1.0} & \textbf{1.0} & \textbf{1.0} & \textbf{1.0} & \textbf{1.1} \\
\hline
\multirow{5}{*}{\rotatebox[origin=c]{90}{kitchen}}
 & MS & .7 & \textbf{.77} & \textbf{.87} & \textbf{.87} & .67 & .67 & .8 & .9 & \textbf{.9} & \textbf{.9} & \textbf{1.2} \\
 & RG09 & .66 & .73 & .77 & .83 & .83 & .83 & .83 & \textbf{.93} & .8 & .8 & 1.4\\
 & LP & .51 & .54 & .58 & .73 & .77 & .77 & .7 & .76 & .67 & .67 & 1.8\\
 & PLR & .33 & .33 & .24 & .16 & .33 & .33 & .33 & .53 & .53 & .53 & 1.5\\
 & FPV(ours) & \textbf{.77} & \textbf{.77} & \textbf{.87} & \textbf{.87} & \textbf{.87} & \textbf{.87} & \textbf{.87} & .87 & .77 & .77 & \textbf{1.2} \\
\hline
\multirow{5}{*}{\rotatebox[origin=c]{90}{logistics}}
 & MS & .35 & .41 & .47 & .6 & .7 & .73 & .76 & \textbf{.89} & \textbf{.97} & \textbf{1.0} & 1.5\\
 & RG09 & .34 & .47 & .58 & \textbf{.69} & .72 & .77 & .8 & .82 & .93 & \textbf{1.0} & 1.9\\
 & LP & .16 & .35 & .41 & .48 & .5 & .53 & .68 & .68 & .69 & .74 & 3.7\\
 & PLR & .23 & .39 & .42 & .51 & .58 & .76 & \textbf{.83} & \textbf{.89} & \textbf{.97} & \textbf{1.0} & \textbf{1.2} \\
 & FPV(ours) & \textbf{.42} & \textbf{.5} & \textbf{.59} & \textbf{.69} & \textbf{.77} & \textbf{.81} & \textbf{.83} & \textbf{.89} & \textbf{.97} & \textbf{1.0} & \textbf{1.2} \\
\hline
\multirow{5}{*}{\rotatebox[origin=c]{90}{miconic}}
 & MS & .41 & .65 & .73 & \textbf{.84} & .88 & .86 & .89 & \textbf{.96} & \textbf{1.0} & \textbf{1.0} & 1.3\\
 & RG09 & .32 & .49 & .59 & .74 & .82 & .87 & .9 & .95 & \textbf{1.0} & \textbf{1.0} & 1.7\\
 & LP & .24 & .43 & .58 & .57 & .76 & .8 & .82 & .86 & .95 & .93 & 2.0\\
 & PLR & .34 & .54 & .64 & .71 & .77 & .86 & .89 & \textbf{.96} & \textbf{1.0} & \textbf{1.0} & \textbf{1.0} \\
 & FPV(ours) & \textbf{.61} & \textbf{.77} & \textbf{.8} & .83 & \textbf{.9} & \textbf{.92} & \textbf{.93} & \textbf{.96} & \textbf{1.0} & \textbf{1.0} & 1.2\\
\hline
\multirow{5}{*}{\rotatebox[origin=c]{90}{rovers}}
 & MS & .46 & \textbf{.6} & .69 & .89 & .89 & \textbf{.96} & \textbf{.96} & \textbf{1.0} & \textbf{1.0} & \textbf{1.0} & 1.2\\
 & RG09 & .43 & \textbf{.6} & .74 & \textbf{.91} & \textbf{.92} & .93 & .93 & .96 & .98 & .98 & 1.5\\
 & LP & .31 & .48 & .64 & .67 & .7 & .74 & .81 & .84 & .91 & .93 & 2.0\\
 & PLR & \textbf{.48} & .52 & \textbf{.8} & .82 & .86 & \textbf{.96} & \textbf{.96} & \textbf{1.0} & \textbf{1.0} & \textbf{1.0} & 1.1\\
 & FPV(ours) & .47 & .58 & .78 & .85 & .89 & \textbf{.96} & \textbf{.96} & \textbf{1.0} & \textbf{1.0} & \textbf{1.0} & \textbf{1.0} \\
\hline
\multirow{5}{*}{\rotatebox[origin=c]{90}{satellite}}
 & MS & .35 & .31 & .36 & .43 & .59 & .66 & .74 & .81 & .81 & .81 & 2.4\\
 & RG09 & .35 & .35 & .42 & .54 & .64 & .62 & .67 & .71 & .74 & .77 & 2.7\\
 & LP & .26 & .29 & .38 & .52 & .52 & .62 & .72 & .71 & .72 & .75 & 2.9\\
 & PLR & .34 & .24 & .29 & .6 & \textbf{.77} & .8 & .82 & .91 & \textbf{.95} & \textbf{.96} & 1.4\\
 & FPV(ours) & \textbf{.48} & \textbf{.47} & \textbf{.58} & \textbf{.68} & .76 & \textbf{.81} & \textbf{.88} & \textbf{.94} & .94 & \textbf{.96} & \textbf{1.1} \\
\hline
\multirow{5}{*}{\rotatebox[origin=c]{90}{sokoban}}
 & MS & \textbf{.28} & \textbf{.3} & .16 & .14 & .27 & .3 & .34 & .25 & .29 & .3 & 1.2\\
 & RG09 & .15 & .22 & .32 & .45 & \textbf{.58} & .6 & .61 & \textbf{.68} & .73 & .8 & 2.7\\
 & LP & .16 & .25 & \textbf{.36} & \textbf{.5} & .57 & \textbf{.63} & \textbf{.68} & .64 & .83 & .87 & 1.9\\
 & PLR & .13 & .21 & .25 & .32 & .5 & .57 & .57 & .5 & \textbf{.86} & \textbf{.96} & \textbf{1.0} \\
 & FPV(ours) & .08 & .24 & .34 & .46 & .55 & .52 & .56 & .61 & .68 & .79 & \textbf{1.0} \\
\hline
\multirow{5}{*}{\rotatebox[origin=c]{90}{zeno-travel}}
 & MS & .32 & .36 & .57 & \textbf{.68} & \textbf{.82} & \textbf{.95} & \textbf{.96} & .98 & \textbf{1.0} & \textbf{1.0} & 1.2\\
 & RG09 & .28 & .29 & .38 & .48 & .59 & .73 & .83 & .89 & .93 & .96 & 1.8\\
 & LP & .22 & .3 & .42 & .52 & .6 & .65 & .85 & .92 & .95 & \textbf{1.0} & 2.4\\
 & PLR & .3 & .41 & .45 & .48 & .73 & .82 & .89 & .96 & \textbf{1.0} & \textbf{1.0} & \textbf{1.0} \\
 & FPV(ours) & \textbf{.41} & \textbf{.47} & \textbf{.58} & .64 & .78 & \textbf{.95} & .95 & \textbf{1.0} & \textbf{1.0} & \textbf{1.0} & \textbf{1.0} \\
\hline
\end{tabular}
}
\end{table}

\subsection{Experimental Results and Discussion}
\label{subsec:ExperimentalResultsDiscussion}
\paragraph{Goal Recognition Performance.}
Table \ref{tab:evaluationTable} shows the experimental results for all evaluated approaches when applied to the 15 benchmark domains.
The average precision is reported for different $\lambda$ values.
As mentioned earlier in this section, we report the averaged precision over 20 repeated runs for FPV.
We also analyzed the standard deviation of FPV's recognition precision over the 20 runs and found that it is between 0 and 0.077, with an average of 0.016 for all benchmark domains.
Hence, as the standard deviation is very low, we omitted it from Table \ref{tab:evaluationTable} for better readability.
Column \textbf{S} reports the average spread (size of the set of goals recognized as the true goal) for each domain.
The larger this value is, the smaller the value of the precision tends to be, as predicting a larger number of goals to be the true goal generally decreases the precision measure (cf., Equation \ref{eq:Precision}).
The results show that the FPV method, on average, achieves the best goal recognition precision for all values of $\lambda$.
Especially when dealing with low observability (i.e., $0.3\leq \lambda \geq 0.9$), FPV significantly outperforms all other approaches.
Nevertheless, it is also interesting to note that although FPV outperforms all other approaches on average, no single approach outperforms all other approaches in all domains.
For example, the MS approach achieves superior performance in the driverlog dataset, the PLR approach achieves superior performance in the dwr dataset.

\paragraph{Computational Efficiency.}
\begin{table}[t!]
\caption{Comparison of average computation times \textbf{in seconds} among all evaluated methods when applied to all evaluated benchmark domains. The average computation time is reported for different sizes of the observation sequence (i.e., $|\mathbf{O}|)$ and set of possible goals (i.e., $|G|$).}
\label{tab:computationTimeTable}
\centering
\resizebox*{0.75\columnwidth}{!}{
\begin{tabular}{ll|lllll}
\multicolumn{1}{c}{$|G|$}		 & \multicolumn{1}{c|}{}		 & \multicolumn{5}{c}{\textbf{$|\mathbf{O}|$}}\\
			 & 			 & 5 		 & 10 		 & 25		 & 50 		 & 100\\ \hline 
\multirow{5}{*}{5}
 & MS & 47 & 87 & 208 & 409 & 811\\
 & RG09 & 14 & 27 & 68 & 136 & 272\\
 & LP & 9 & 18 & 44 & 87 & 175\\
 & PLR & 1.4 & 1.4 & 1.4 & 1.4 & 1.4\\
 & FPV(ours) & \textbf{.37} & \textbf{.37} & \textbf{.37} & \textbf{.37} & \textbf{.37}\\
\hline
\multirow{5}{*}{10}
 & MS & 94 & 174 & 416 & 818 & 1622\\
 & RG09 & 28 & 54 & 136 & 272 & 544\\
 & LP & 18 & 35 & 87 & 175 & 349\\
 & PLR & 2.8 & 2.8 & 2.8 & 2.9 & 2.9\\
 & FPV(ours) & \textbf{.74} & \textbf{.74} & \textbf{.74} & \textbf{.74} & \textbf{.74}\\
\hline
\multirow{5}{*}{20}
 & MS & 188 & 348 & 832 & 1636 & 3244\\
 & RG09 & 56 & 108 & 272 & 544 & 1088 \\
 & LP & 35 & 70 & 175 & 349 & 698 \\
 & PLR & 5.7 & 5.7 & 5.7 & 5.7 & 5.7 \\
 & FPV(ours) & \textbf{1.48} & \textbf{1.48} & \textbf{1.48} & \textbf{1.48} & \textbf{1.48} \\
\hline
\end{tabular}
}
\end{table}
Table \ref{tab:computationTimeTable} reports the extrapolated average cumulated computation time for all evaluated approaches, averaged over all 15 benchmark domains.
The computation time is reported \textit{in seconds} for different potential sizes of observation sequences (i.e., $|O_t|$) and sets of possible goals (i.e., $|G|$).
The values in this table are calculated from average computation time values per observation and goal for each approach and are averaged over all experiments.
The table's primary purpose is to visualize how well the different recognition approaches scale regarding computation time when the size of the considered goal recognition problem is altered.
The results show that the FPV approach is the most computationally efficient among all evaluated approaches.
Table \ref{tab:computationTimeTable} visualizes the advantages of FPV regarding computational efficiency compared to the other evaluated approaches very well, especially when the size of the goal recognition problems is scaled.
One can see that the computation time that FPV requires does not increase with an increasing number of observations.
This is because once FPV has determined fact observation probabilities for each goal, FPV only has to compute the heuristic values for each goal for each new observation.
The same can be observed for the PLR approach for a similar reason as for FPV.
PLR only has to compute some heuristics based on the landmarks extracted upfront.
The main computation time that PLR requires is due to the landmark extraction process, which is required once per online goal recognition problem, similar to the fact observation probability estimation of FPV.
Nevertheless, the results show that FPV's extraction process is much more efficient than that of PLR, which results in FPV being roughly four times faster than PLR, the second fastest evaluated approach.
As expected, the MS approach is the least computationally efficient, as it requires computing an entire non-relaxed plan per goal per new observation step.
Interestingly, although RG09 uses relaxed plans, which can be computed more efficiently than non-relaxed plans, it also does not scale well with increasing observations.
The main reason for this is most probably that the RG09 approach also recomputes a relaxed plan for each goal for each new observation step.
In addition, it requires the planner to include all observations already observed in the computed plan.
This results in a much more complex planning problem.

The results also show that when $|G|$ increases, the FPV and PLR approaches scale less well than when $|O|$ increases.
This is because both FPV and PLR require the most computation time to compute fact observation probabilities or landmarks.
With an increasing number of potential goals, this process, of course, requires more time.
Nevertheless, one can see that the FPV approach still scales much better than the PLR approach and also much better than the MS, LP and RG09 approaches, for which the scaling is even worse than for PLR.

\section{Related Work}
\label{sec:relatedWork}
Since the idea of Plan Recognition as Planning was introduced by \citet{ramirez2009plan}, many approaches have adopted this paradigm \cite{ramirez2010probabilistic, ramirez2011goal, yolanda2015fast, vered2016mirroring, sohrabi2016revisited, masters2017cost, pereira2020landmark, cohausz2022plan}.
It was recognized relatively soon that the initial PRAP approaches are computationally demanding, as they require computing entire plans.
Since then, this problem has been addressed by many studies with the approach by \citet{pereira2020landmark} being a recent example.
This method also belongs to a recent type of PRAP method (to which FPV also belongs), which does not derive probability distributions over the set of possible goals by analyzing cost differences but ranks the possible goals by calculating heuristic values.
Additional approaches from this area include a variant that was suggested as an approximation for their main approach by \citet{ramirez2009plan} and the Linear Programming approach, which was introduced by Santos et al. \cite{santos2021lp}.

\section{Conclusion}
\label{sec:conclusion}
In conclusion, we presented FPV, a new goal recognition method based on comparing fact observation probabilities with the set of actually observed facts.
We empirically evaluated the FPV and showed that it achieves better goal recognition precision than four state-of-the-art goal recognition methods, especially early in an observation sequence.
FPV is also four times more efficient regarding computation time than the second most efficient approach, PLR, and roughly 2000 times faster than the least efficient approach, MS.
FPV also scales much better in terms of observation sequence length compared to most of the evaluated approaches.
FPV's advantage regarding computational efficiency is mainly due to the fact that FPV computes the fact observation probabilities offline, which leaves very little computational workload during online recognition.
In future work, it would be interesting to consider evaluation scenarios based on actual human behavior.
Another exciting path would be exploring different approaches to estimating the fact observation probabilities.




\begin{ack}
This work was funded by the German Federal Ministry for the Environment, Nature Conservation, Nuclear Safety and Consumer Protection (Research Grant: 67WM22003, Project: GreenPickUp) and the German Federal Ministry for Economic Affairs and Climate Action (Research Grant: 01ME23002, Project: MEDICAR 4.0).
\end{ack}



\bibliography{ecai24}

\begin{thebibliography}{20}
\providecommand{\natexlab}[1]{#1}
\providecommand{\url}[1]{\texttt{#1}}
\expandafter\ifx\csname urlstyle\endcsname\relax
  \providecommand{\doi}[1]{doi: #1}\else
  \providecommand{\doi}{doi: \begingroup \urlstyle{rm}\Url}\fi

\bibitem[Amado et~al.(2018)Amado, Aires, Pereira, Magnaguagno, Granada, and Meneguzzi]{amado2018lstm}
L.~Amado, J.~P. Aires, R.~F. Pereira, M.~C. Magnaguagno, R.~Granada, and F.~Meneguzzi.
\newblock Lstm-based goal recognition in latent space.
\newblock \emph{arXiv preprint arXiv:1808.05249}, 2018.

\bibitem[Amado et~al.(2023)Amado, Pereira, and Meneguzzi]{amado2023robust}
L.~R. Amado, R.~F. Pereira, and F.~Meneguzzi.
\newblock Robust neuro-symbolic goal and plan recognition.
\newblock In \emph{Proceedings of the 37th AAAI Conference on Artificial Intelligence (AAAI), 2023, Estados Unidos.}, 2023.

\bibitem[Bonet et~al.(1997)Bonet, Loerincs, and Geffner]{bonet1997robust}
B.~Bonet, G.~Loerincs, and H.~Geffner.
\newblock A robust and fast action selection mechanism for planning.
\newblock In \emph{AAAI/IAAI}, pages 714--719, 1997.

\bibitem[Cohausz et~al.(2022)Cohausz, Wilken, and Stuckenschmidt]{cohausz2022plan}
L.~Cohausz, N.~Wilken, and H.~Stuckenschmidt.
\newblock Plan-similarity based heuristics for goal recognition.
\newblock In \emph{2022 IEEE International Conference on Pervasive Computing and Communications Workshops and other Affiliated Events (PerCom Workshops)}, pages 316--321. IEEE, 2022.

\bibitem[Geib(2002)]{geib2002problems}
C.~W. Geib.
\newblock Problems with intent recognition for elder care.
\newblock In \emph{Proceedings of the AAAI-02 Workshop “Automation as Caregiver}, pages 13--17, 2002.

\bibitem[Geib and Goldman(2001)]{geib2001plan}
C.~W. Geib and R.~P. Goldman.
\newblock Plan recognition in intrusion detection systems.
\newblock In \emph{Proceedings DARPA Information Survivability Conference and Exposition II. DISCEX'01}, volume~1, pages 46--55. IEEE, 2001.

\bibitem[Hoffmann(2003)]{hoffmann2003metric}
J.~Hoffmann.
\newblock The {M}etric-{FF} planning system: Translating ``ignoring delete lists'' to numeric state variables.
\newblock \emph{JAIR}, 20:\penalty0 291--341, 2003.

\bibitem[Hoffmann et~al.(2004)Hoffmann, Porteous, and Sebastia]{hoffmann2004ordered}
J.~Hoffmann, J.~Porteous, and L.~Sebastia.
\newblock Ordered landmarks in planning.
\newblock \emph{Journal of Artificial Intelligence Research}, 22:\penalty0 215--278, 2004.

\bibitem[Masters and Sardina(2017)]{masters2017cost}
P.~Masters and S.~Sardina.
\newblock Cost-based goal recognition for path-planning.
\newblock In \emph{Proceedings of the 16th Conference on Autonomous Agents and MultiAgent Systems}, pages 750--758, 2017.

\bibitem[McDermott et~al.(1998)McDermott, Ghallab, Howe, Knoblock, Ram, Veloso, Weld, and Wilkins]{mcdermott1998pddl}
D.~McDermott, M.~Ghallab, A.~Howe, C.~Knoblock, A.~Ram, M.~Veloso, D.~Weld, and D.~Wilkins.
\newblock Pddl-the planning domain definition language, 1998.

\bibitem[Pereira et~al.(2020)Pereira, Oren, and Meneguzzi]{pereira2020landmark}
R.~F. Pereira, N.~Oren, and F.~Meneguzzi.
\newblock Landmark-based approaches for goal recognition as planning.
\newblock \emph{Artificial Intelligence}, 279:\penalty0 103217, 2020.

\bibitem[Pynadath and Wellman(1995)]{pynadath1995accounting}
D.~V. Pynadath and M.~P. Wellman.
\newblock Accounting for context in plan recognition, with application to traffic monitoring.
\newblock In \emph{Proceedings of the Eleventh conference on Uncertainty in artificial intelligence}, pages 472--481, 1995.

\bibitem[Ram\'{\i}rez and Geffner(2009)]{ramirez2009plan}
M.~Ram\'{\i}rez and H.~Geffner.
\newblock Plan recognition as planning.
\newblock In \emph{Proceedings of the 21st International Joint Conference on Artificial Intelligence}, IJCAI'09, page 1778–1783, 2009.

\bibitem[Ram\'{\i}rez and Geffner(2010)]{ramirez2010probabilistic}
M.~Ram\'{\i}rez and H.~Geffner.
\newblock Probabilistic plan recognition using off-the-shelf classical planners.
\newblock In \emph{Proceedings of the Twenty-Fourth AAAI Conference on Artificial Intelligence}, AAAI'10, page 1121–1126. AAAI Press, 2010.

\bibitem[Ramirez and Geffner(2011)]{ramirez2011goal}
M.~Ramirez and H.~Geffner.
\newblock Goal recognition over pomdps: Inferring the intention of a pomdp agent.
\newblock In \emph{Twenty-second international joint conference on artificial intelligence}, 2011.

\bibitem[Santos et~al.(2021)Santos, Meneguzzi, Pereira, and Pereira]{santos2021lp}
L.~R. Santos, F.~Meneguzzi, R.~F. Pereira, and A.~G. Pereira.
\newblock An lp-based approach for goal recognition as planning.
\newblock In \emph{Proceedings of the AAAI Conference on Artificial Intelligence}, volume~35, pages 11939--11946, 2021.

\bibitem[Sohrabi et~al.(2016)Sohrabi, Riabov, and Udrea]{sohrabi2016revisited}
S.~Sohrabi, A.~V. Riabov, and O.~Udrea.
\newblock Plan recognition as planning revisited.
\newblock In \emph{Proceedings of the Twenty-Fifth International Joint Conference on Artificial Intelligence}, IJCAI'16, page 3258–3264. AAAI Press, 2016.
\newblock ISBN 9781577357704.

\bibitem[Vered et~al.(2016)Vered, Kaminka, and Biham]{vered2016mirroring}
M.~Vered, G.~Kaminka, and S.~Biham.
\newblock Online goal recognition through mirroring: humans and agents.
\newblock In K.~Forbus, T.~Hinrichs, and C.~Ost, editors, \emph{Fourth Annual Conference on Advances in Cognitive Systems}, Advances in Cognitive Systems. Cognitive Systems Foundation, 2016.
\newblock URL \url{http://www.cogsys.org/2016}.
\newblock Annual Conference on Advances in Cognitive Systems 2016 ; Conference date: 23-06-2016 Through 26-06-2016.

\bibitem[Wilken and Stuckenschmidt(2021)]{wilken2021hybrid}
N.~Wilken and H.~Stuckenschmidt.
\newblock Combining symbolic and statistical knowledge for goal recognition in smart home environments.
\newblock In \emph{2021 IEEE International Conference on Pervasive Computing and Communications Workshops and other Affiliated Events (PerCom Workshops)}, pages 26--31, 2021.
\newblock \doi{10.1109/PerComWorkshops51409.2021.9431145}.

\bibitem[Yolanda et~al.(2015)Yolanda, R-Moreno, Smith, et~al.]{yolanda2015fast}
E.~Yolanda, M.~D. R-Moreno, D.~E. Smith, et~al.
\newblock A fast goal recognition technique based on interaction estimates.
\newblock In \emph{Twenty-Fourth International Joint Conference on Artificial Intelligence}, 2015.

\end{thebibliography}

\end{document}